\documentclass[10pt,twocolumn,letterpaper]{article}

\usepackage[pagenumbers]{cvpr} %

\usepackage[dvipsnames]{xcolor}

\usepackage{epsfig}
\usepackage{graphicx}
\usepackage{comment}
\usepackage{amsmath,amssymb} %
\usepackage{xcolor}
\usepackage[export]{adjustbox}
\usepackage{bbm}

\usepackage[font=small,skip=5pt]{caption}
\usepackage{float}

\usepackage{multirow}
\usepackage{tabularx}
\usepackage{soul}
\usepackage{array}
\usepackage{comment}
\usepackage{outlines}

\usepackage[mathscr]{euscript}
\newcolumntype{L}[1]{>{\raggedright\let\newline\\arraybackslash\hspace{0pt}}m{#1}}
\newcolumntype{C}[1]{>{\centering\let\newline\\arraybackslash\hspace{0pt}}m{#1}}
\newcolumntype{R}[1]{>{\raggedleft\let\newline\\arraybackslash\hspace{0pt}}m{#1}}
\newcolumntype{Y}{>{\centering\arraybackslash}X}
\usepackage{booktabs}
\usepackage{enumitem}

\usepackage{amsmath}

\newcommand{\dataName}{ProciGen-Video}
\newcommand{\dataNameShort}{ProciGen-V} %
\newcommand{\methodName}{InterTrack} %
\newcommand{\humAEName}{CorrAE} %
\newcommand{\poseName}{TOPNet} %

\newcommand{\mat}[1]{\mathbf{#1}}

\newcommand{\vect}[1]{\mathbf{#1}}

\newcommand{\pose}[0]{\boldsymbol{\theta}}
\newcommand{\shape}[0]{\boldsymbol{\beta}}

\definecolor{ForestGreen}{RGB}{14,109,14}

\definecolor{cvprblue}{rgb}{0.21,0.49,0.74}
\usepackage[pagebackref,breaklinks,colorlinks,citecolor=cvprblue]{hyperref}

\def\paperID{66} %
\def\confName{3DV\xspace}
\def\confYear{2025\xspace}
\begin{document}

\twocolumn[{%
\renewcommand\twocolumn[1][]{#1}%

\title{\methodName{}: Tracking Human Object Interaction without Object Templates}
\author{
Xianghui Xie$^{1,2,3}$ 
\qquad
Jan Eric Lenssen$^3$ 
\qquad
Gerard Pons-Moll$^{1,2,3}$ \\
\\
{\small $^1$University of T\"ubingen, Germany \hspace{1cm} $^2$T\"ubingen AI Center, Germany  } \\
{\small $^3$Max Planck Institute for Informatics, Saarland Informatic Campus, Germany\hspace{1cm}} \\
{\small\href{https://virtualhumans.mpi-inf.mpg.de/InterTrack/}{https://virtualhumans.mpi-inf.mpg.de/InterTrack/}}\\
}

\maketitle
\begin{center}
    \centering
    \captionsetup{type=figure}
    \includegraphics[width=1.0\textwidth]{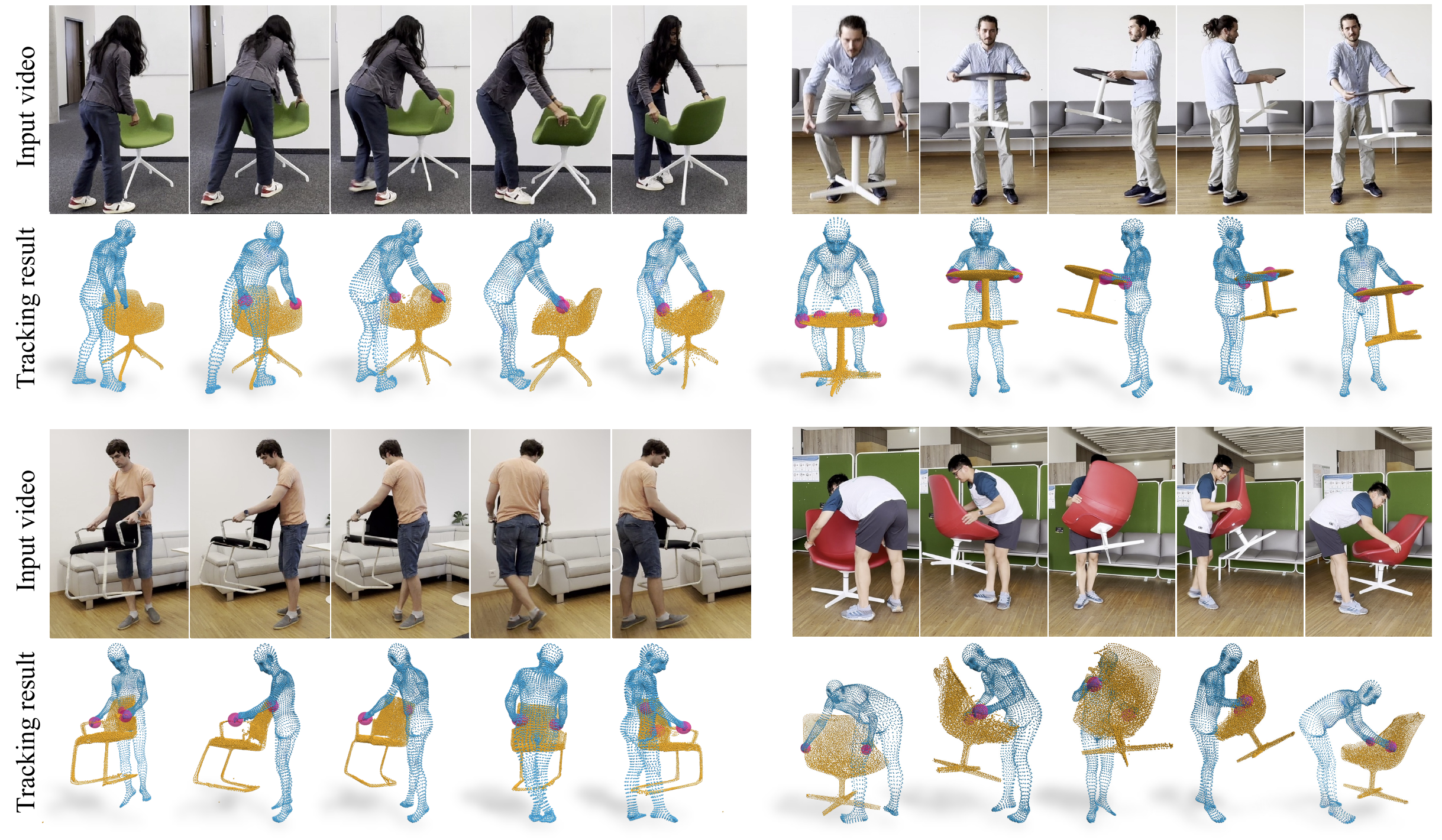}
    \captionof{figure}{From a monocular RGB video, our method tracks the human and object under occlusion and dynamic motions, without using any object templates. Our method is trained \emph{only} on synthetic data and generalizes well to real-world videos captured by mobile phones.}
    \label{fig:teaser}
\end{center}%
}]

\begin{abstract}
Tracking human object interaction from videos is important to understand human behavior from the rapidly growing stream of video data. Previous video-based methods require predefined object templates while single-image-based methods are template-free but lack temporal consistency. In this paper, we present a method to track human object interaction without any object shape templates. We decompose the 4D tracking problem into per-frame pose tracking and canonical shape optimization. We first apply a single-view reconstruction method to obtain temporally-inconsistent per-frame interaction reconstructions. 
Then, for the human, we propose an efficient autoencoder to predict SMPL vertices directly from the per-frame reconstructions, introducing temporally consistent correspondence. 
For the object, we introduce a pose estimator that leverages temporal information to predict smooth object rotations under occlusions. To train our model, we propose a method to generate synthetic interaction videos and synthesize in total 10 hour videos of 8.5k sequences with full 3D ground truth. Experiments on BEHAVE and InterCap show that our method significantly outperforms previous template-based video tracking and single-frame reconstruction methods. Our proposed synthetic video dataset also allows training video-based methods that generalize to real-world videos. Our code and dataset will be publicly released.

\end{abstract}    
\section{Introduction}
\label{sec:intro}

Jointly reconstructing humans and objects is an important task to understand humans and their interaction with the environment. In this paper, we address the problem of tracking human object interaction from a single RGB camera, without any object templates. This is a very challenging task due to depth scale ambiguity, heavy occlusions and dynamic human object motions. Moreover, the human and object poses as well as shapes need to be estimated simultaneously as no prior templates are given. 

Earlier work VisTracker~\cite{xie2023vistracker} pioneered monocular interaction tracking by reasoning about the occluded object using visible frames and human information. However, they rely on pre-defined object templates which limits its applicability to general scenarios. Furthermore, they use per-frame object pose estimator with ad-hoc smoothing to reason occluded objects, which does not fully explore temporal information. 
More recently, HDM~\cite{xie2023template_free} proposed a template-free approach to reconstruct human and object from single images. Their model trained only on synthetic data shows strong generalization ability to real-world images. However, this method fails under heavy occlusions and produces inconsistent shapes across frames.

To address these challenges, we propose \textbf{\methodName{}}, interaction tracking without object templates. We decompose the 4D video reconstruction problem into per-frame pose estimation and global shape optimization. This decomposition greatly constraints the solution space of 4D reconstruction, making the problem tractable. Specifically, we start with the HDM~\cite{xie2023template_free} per-frame reconstructions which provide initial 3D human and object point clouds that are inconsistent across frames. For the human, we propose a simple and efficient autoencoder, \textbf{\humAEName{}}, which directly predicts the SMPL~\cite{smpl2015loper} vertices that are aligned with the HDM human reconstructions. This allows us to obtain disentangled SMPL pose and shape parameters for temporally consistent optimization and introduces correspondence over time. For the object, we introduce \textbf{\poseName{}}, which leverages temporal information to predict object rotation from monocular RGB video. The temporal design allows accurate object pose prediction even under heavy occlusions. With the predicted rotations, we can then optimize the object shape in canonical space as well as per-frame pose transformations, leading to temporally consistent tracking. Last, we jointly optimize human and object based on the predicted contacts, leading to more plausible interactions. 

Prior works~\cite{xie22chore, xie2023vistracker} train their methods on real data and test on same set of object instances, which limits the generalization to new objects~\cite{xie2023template_free}. Recent work ProciGen~\cite{xie2023template_free} proposed a synthetic interaction dataset with 1M images and 21k different object shapes. However, ProciGen contains only static frames, making it impossible to train video-based methods. To this end, we propose \textbf{\dataName{}}, a method to generate synthetic interaction videos and we generate 8.5k videos of 10 categories paired with full 3D ground truth. This dataset allows us to train our object pose estimator \poseName{} that generalizes to real videos. 

We evaluate our method on BEHAVE~\cite{bhatnagar22behave} and InterCap~\cite{huang2022intercap} dataset. Experiments show that our method significantly outperforms VisTracker~\cite{xie2023vistracker} (which requires template) and HDM~\cite{xie2023template_free}. Our ablation shows that our \humAEName{} achieves similar performance compared to SoTA human registration method NICP~\cite{marin24nicp} but is 30 times faster, and our proposed \poseName{} works significantly better than prior category-level pose estimator CenterPose~\cite{lin2022icra:centerpose}. Results also show that pre-training on our \dataName{} dataset helps boost the performance and our model trained on synthetic \dataName{} generalizes to real videos. 

In summary, our key contributions are: 
\begin{itemize}
    \item We propose \methodName{}, the first method to track full-body dynamic object interaction from monocular RGB videos without object templates. 
    \item We introduce an efficient autoencoder for human registration that is 30 times faster with comparable performance. 
    \item We propose a video-based object pose estimator that leverages temporal information to predict object rotations even under heavy occlusions. 
    \item We introduce \dataName{}, a method to generate synthetic videos for interaction. With this, we create a dataset of 8.5k videos paired with full 3D ground truth. 
\end{itemize}

\section{Related Works}
\textbf{General object tracking.} Reconstructing and tracking objects from monocular videos has been studied for decades~\cite{yunus2024nonrigidstar}. Early works~\cite{newcombe2011kinectfusion, newcombe2015dynamicfusion} track objects with TSDF fusion and are further improved with neural networks ~\cite{wen2023bundlesdf, xue2023nsf}. They require depth which can be inconvenient hence more recent works track human \cite{VIBE:CVPR:2020, weng_humannerf_2022_cvpr, jiang2024multiply} or learn general articulated objects~\cite{yang2021lasr, yang2021viser} from RGB videos. Follow up works extend to learn shapes from casual videos~\cite{yang2022banmo, yang2023rac} or unstructured images in the wild~\cite{wu2023magicpony, li2024_3DFauna}. These methods learn an explicit template shape, the deformation skeleton and skinning weights. Orthogonal to this, NPG~\cite{das2023npg} and KeyTr~\cite{CVPR22keytr} represent shapes as basis points and coefficients learned from videos. Despite impressive results, these methods assume single object and cannot handle the compositional shape during human object interaction.

\noindent\textbf{Interaction reconstruction and tracking.} Interaction modelling has recently received more and more attention in generation~\cite{petrov2020objectpopup, zhang2022couch, li2023omomo, zhou2024gears, li2024genzi} and reconstruction~\cite{xie22chore, hasson19_obman, zhou2022toch, xie2024rhobin, zhang2024hoi_m3, xue2024human3diffusion}. Hand-object reconstruction has been well studied and recent methods are able to recover interaction without any templates~\cite{ye2023ghop, ye2023affordance, qiHOISDFConstrainHand3D2024}. Full body object interaction methods usually rely on template objects~\cite{zhang2020phosa, xie22chore, wang2022reconstruction, nam2024contho, weng2020holistic, chenKinematicsbased3DHOI2024, huoStackFLOWHOI2023}, with only one template-free method \cite{xie2023template_free} trained on large synthetic dataset. 3D interaction from single images is heavily ill-posed hence some methods leverage temporal information to improve robustness, with works that track interaction from multi-view\cite{bhatnagar22behave, jiang2022neuralfusion, huang2022rich, huang2022intercap, zhang2023neuraldome} or monocular~\cite{robustfusion-arxiv, jiang2023instantnvr} RGBD cameras. From RGB camera only, researchers leverage photometric consistency~\cite{hasson20_handobjectconsist} or large diffusion model~\cite{ye2023vhoi} to guide the reconstruction and lots of methods can obtain good object shapes~\cite{ye2022hand_object, fan2024hold, hampali2023inhand, huang2022hhor, shiFreeMovingHandObject2024} from hand-object interaction. Object shape is less constrained by the human during full-body interaction hence methods rely on pre-scanned point clouds~\cite{CVPR21HPS, guzov24ireplica} or meshes~\cite{xie2023vistracker, xu2021d3dhoi} to track the interaction. Despite being robust to occlusion, the reliance on object templates limits their applicability. In contrast, our method deals with full-body dynamic object interaction and does not require any object templates.

\noindent\textbf{Correspondence estimation.} Correspondence is the key for video tracking. For 3D humans, registering common templates like SMPL~\cite{smpl2015loper} to 3D scans is a classic problem and has been studied in many works~\cite{groueix2018b_3DCODED, marin2018farm_FM_registration, bhatnagar2020loopreg, marin24nicp, bhatnagar2020ipnet, coronaLVD, feng2023arteq}. Most SoTA methods~\cite{bhatnagar2020loopreg, bhatnagar2020ipnet, marin24nicp, coronaLVD} rely on optimization, which is slow for video processing. In contrast, our \humAEName{} directly predicts SMPL vertices which is more efficient. 

\noindent For 3D objects, correspondence can be obtained via functional maps~\cite{deepFuctionalMaps2017iccv, functionalmaps2012Maks} or deep neural networks~\cite{zhou2022art, das2023npg, wewer2023simnp, schroppel23npcd, halimi2019unsupervised_corr, zheng2021dit}. However, learning-based methods can only process shapes with aligned orientation which does not apply to our objects that have arbitrary rotations during interaction. Instead, we first estimate the rotations and then optimize the shape in canonical space.

\begin{figure*}
    \centering
    \includegraphics[width=1.0\linewidth]{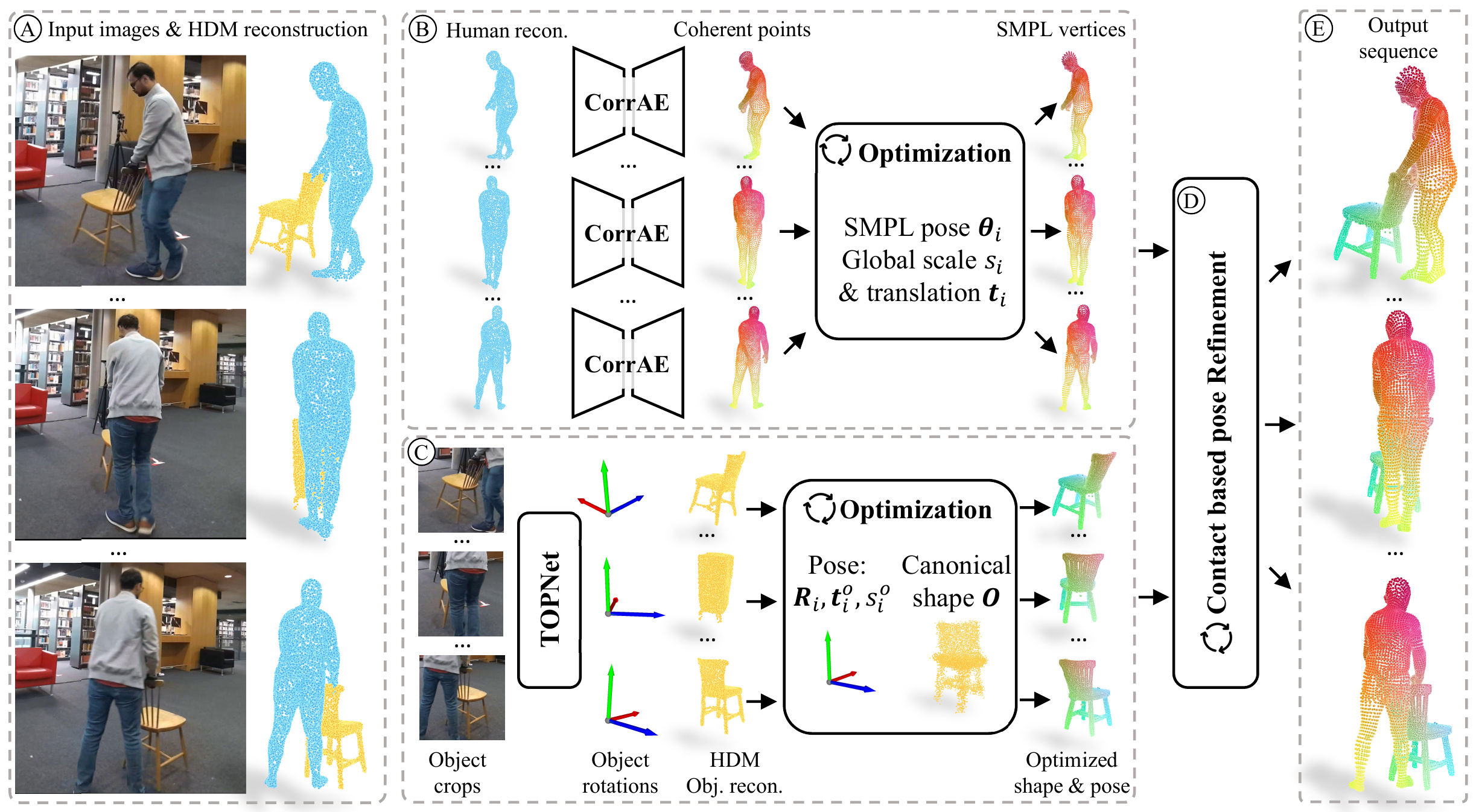}
    \caption{\textbf{Method Overview.} Given an image sequence of human object interaction and HDM~\cite{xie2023template_free} reconstructions (A), we aim at obtaining coherent tracking of the human and object across frames (E). We first use a simple yet efficient autoencoder \humAEName{} to obtain coherent humans points and optimize human via the SMPL layer (B, \cref{subsec:human-recon}). We then use a temporal object pose estimator \poseName{} to predict the object rotation, which allows us to optimize a common object shape in canonical space and fine tune pose predictions (C, \cref{subsec:object-recon}). We then jointly optimize human and object based on contacts to obtain consistent tracking (D, \cref{subsec:joint-track}). }
    \label{fig:method-overview}
\end{figure*}

\section{Method}\label{sec:method}

We present \methodName{}, an approach to track interacting human and object from monocular video, without object shape templates. This is very challenging as one needs to establish correspondences, and to reason about object pose/shape simultaneously under heavy occlusion and dynamic motion. 

Our key idea is to decompose 4D human and object into per-frame poses and global consistent shapes. For the human, we propose a novel autoencoder that directly predicts SMPL~\cite{smpl2015loper} vertices from unordered points, allowing us to use the disentangled SMPL pose and shape parameters. For the object, we introduce a video based object pose predictor that estimates temporally consistent object rotations. This enables us to optimize one common shape in canonical space and per-frame object transformations. An overview of our method can be found in \cref{fig:method-overview}. 

In this section, we first briefly discuss how do we obtain per-frame 3D reconstruction with HDM~\cite{xie2023template_free} in \cref{subsec:preliminary}. We then introduce our novel \humAEName{} for human correspondence and optimization in \cref{subsec:human-recon} (\cref{fig:method-overview}B), and \poseName{} for temporal rotation prediction and shape optimization of objects in \cref{subsec:object-recon}(\cref{fig:method-overview}C). We then jointly optimize human and object based contacts to obtain plausible interaction (\cref{fig:method-overview}D, \cref{subsec:joint-track}). Prior datasets either have only limited objects~\cite{bhatnagar22behave, huang2022intercap} or static frames~\cite{xie2023template_free}. To train our video pose estimator, we introduce a method to generate synthetic video dataset for human object interaction (\cref{subsec:procigen-video}).

\subsection{Preliminaries}\label{subsec:preliminary}

Given an image sequence $\{\mat{I}_i, i=1,...,T\}$ from monocular RGB camera of a human interacting with an object, we aim at reconstructing temporally consistent 3D human $\mat{H}_i$ and object $\mat{O}_i$ for each image. We represent each human and object as dense point clouds and obtain a sequence of temporally coherent points where each point has correspondence across frames. 

We first apply HDM~\cite{xie2023template_free}, a SoTA template-free approach to reconstruct interacting human and object for each image. 
Specifically, given a single RGB image $\mat{I}_i$, HDM~\cite{xie2023template_free} reconstructs 3D point $\mat{P}^h_i\in\mathbb{R}^{N\times 3}, \mat{P}^o_i\in\mathbb{R}^{N\times 3}$ of human and object respectively. They adopt a conditional generation paradigm which iteratively denoises Gaussian point clouds into clean human and object points, conditioned on the pixel aligned image features for each points. %

The HDM reconstruction provides strong prior of the human-object shapes and interaction semantics like contact points. However, the reconstructed points do not have correspondence across frames due to the nature of the stochastic diffusion process. Furthermore, the diffusion model outputs point clouds in a normalized metric space, leading to different sizes of human and object even if they are from the same sequence. We address these problems next. 

\subsection{Consistent human reconstruction}\label{subsec:human-recon}
The goal is to leverage the shape priors from the HDM predicted points that are \emph{unordered} and create a sequence of points with cross-frame consistency. 
Several previous methods can register the SMPL model~\cite{smpl2015loper} into human point clouds~\cite{bhatnagar2020loopreg, bhatnagar2020ipnet, marin24nicp}. However, these methods use slow optimization which makes it difficult to apply to video data. We propose to use a simple autoencoder, \humAEName{} $f_\text{ae}: \mathbb{R}^{N\times 3}\to \mathbb{R}^{N_s\times 3}$, which directly outputs the SMPL vertices from unordered points. 

Specifically, we use PVCNN~\cite{liu2019pvcnn} as the point cloud encoder which encodes the human points $\mat{P}^h_i$ into a 1D latent vector $\vect{z}_i^h$. We then stack several MLP layers to decode the latent vector into 3D points $\hat{\mat{P}}^h_i\in \mathbb{R}^{3N_s}$, here $N_s=6890$ is the number of vertices in the SMPL model~\cite{smpl2015loper}. Thanks to the regularity of MLP layers, the outputs $\hat{\mat{P}}^h_i$ are ordered points~\cite{xie2023template_free, zhou2022art, wewer2023simnp} and establish correspondence between different human point clouds. Different from~\cite{wewer2023simnp, xie2023template_free} that require aligned object shapes, the key to handle humans with different orientations is to train the network using Chamfer distance combined with vertex to vertex error:
\begin{equation}
    \mathcal{L}_\text{ae} =  d_\text{cd}(f_\text{ae}(\mat{P}), \mat{P}) + \lambda_\text{v2v} ||f_\text{ae}(\mat{P}) - \mat{V}_\text{SMPL}||^2_2
    \label{eq:loss_autoencoder}
\end{equation}
where $\mat{V}_\text{SMPL}, \mat{P}$ are the vertices and surface samples from SMPL meshes respectively. We train this model on the ground truth SMPL meshes from synthetic ProciGen \cite{xie2023template_free} dataset and find it works well on the reconstructed point clouds from HDM~\cite{xie2023template_free}. More importantly, it is much faster than traditional optimization-based method~\cite{marin24nicp}, see \cref{tab:ablation-human-ae}.

With our \humAEName{}, one can establish correspondence across frames and then optimize the human representation for human tracking. One straightforward option is to optimize the latent code $\mat{z}^h_i$ of each frame with temporal smoothness loss. However, the latent space is not interpretable, without decoupling of the human pose and shape. Instead, we leverage the SMPL representation~\cite{smpl2015loper} which has disentangled pose and shape parameter space. 

Specifically, we first use the smplx library~\cite{SMPL-X:2019} to obtain each frame's SMPL parameters $\pose_i, \shape_i$ from our \humAEName{} predictions $\hat{\mat{P}}^h_i$. Note that our \humAEName{} is the key here as without correspondence, it is impossible to obtain accurate SMPL parameters~\cite{bhatnagar2020loopreg, bhatnagar2020ipnet}.
We then compute a mean body shape $\Bar{\shape}=\frac{1}{T}\sum_i\shape_i$ as the shape parameter for the full sequence and optimize only the per-frame translation $\vect{t}_i$, scale $s_i$ and local poses $\pose_i$. Let $\mat{H}(\pose, \shape)$ denote the SMPL model that outputs SMPL vertices given pose and shape parameters, we compute the human points in camera space by: $\mat{H}_i = s_i \mat{H}(\pose_i, \Bar{\shape}) + \vect{t}_i$. We optimize the per-frame parameters $\pose_i, \vect{t}_i, s_i$ to fit into original HDM human reconstructions $\mat{P}^h_i$ with temporal smoothness and pose regularization: 
\begin{equation}
    \mathcal{L}_\text{hum} = \sum_i^T \lambda_\text{cd}^hd_\text{cd}(\mat{P}^h_i, \mat{H}_i) + \lambda_p L_\text{pr}(\pose_i) + \lambda_a^h L_\text{acc}(\mathcal{H}) 
    \label{eq:loss_human}
\end{equation}
here $L_\text{pr}$ is the pose prior loss~\cite{bhatnagar2020ipnet, he24nrdf, tiwari22posendf}, and $L_\text{acc}(\mathcal{H}) = \sum_{i=2}^T||\mat{H}_{i} - 2\mat{H}_{i-1} + \mat{H}_{i-2}||_2^2$ is a smoothness loss based on acceleration. We show in \cref{tab:ablation-human-ae} that optimizing via SMPL model is better than optimizing the latent vectors directly.

\subsection{Rigid object shape reconstruction}\label{subsec:object-recon}
Similar to the human points, the object points $\mat{P}^o_i$ predicted by HDM~\cite{xie2023template_free} are also unordered and lack temporal consistency. Furthermore, the model generates different 3D shapes when the object is occluded in the image, making tracking more challenging. In a video, the object can be decomposed into one global shape in canonical space and per-frame relative transformations. Estimating object pose without shape is however non-trivial as the object can be fully occluded during interaction. Methods like COLMAP~\cite{schoenberger2016sfm_COLMAP} and OnePose~\cite{sun2022onepose} fail to track the object due to limited object features. To this end, we propose \poseName, a transformer-based network that leverages the temporal information to predict object rotations in a video. Our idea is to use transformer~\cite{NIPS2017_attention} to exchange features across frames and output smooth object poses.

Specifically, we first use DINOv2 encoder~\cite{oquab2024dinov2} followed with two additional convolution layers~\cite{wu2023magicpony} to compress each input image $\mat{I}_i$ into a feature vector $\vect{F}^I_i$. We concatenate the image feature with human pose information $\mat{F}^h_i$ and object visibility ratio $v_i\in [0, 1]$ (0-fully occluded and 1-fully visible). The human feature $\mat{F}^h_i=(\pose_i, \mat{J}_i, \mat{J}_i^\prime)$ consists of SMPL pose $\pose_i$, 3D joint location $\mat{J}_i$ and joint velocity $\mat{J}^\prime_i$.  Hence, the feature vector for each image is $\mat{F}_i = (\mat{F}_i^I, \mat{F}_i^h, v_i)$. 
We then stack features from $W$ consecutive frames as a feature matrix $\mathcal{F}=(\mat{F}_1,...,\mat{F}_W)$ and use transformer encoder~\cite{NIPS2017_attention} to exchange temporal information. The attention layer in transformer outputs more temporally consistent features which are sent to MLP to predict object rotation, represented via the 6D representation~\cite{Zhou2019CVPR_obj6d}. 
We train our \poseName{} with L1 distance between the predicted rotations $\hat{\mathcal{R}}=\{\hat{\mat{R}}_1,...,\hat{\mat{R}}_W\} $ and ground truth $\mat{R}_i$, and acceleration loss $L_\text{acc}$ (\cref{eq:loss_human}): $\mathcal{L}_\text{rot} = \sum_i||\hat{\mat{R}}_i - \mat{R}_i||_1 + \lambda^p_a L_\text{acc} (\hat{\mathcal{R}})$. Our \poseName{} is trained with W=16 consecutive frames to learn temporal correlations due to limited data IO speed. At test time, we find it better to use a longer window W=64 and average predictions of each frame in different sliding windows. See Supp. for analysis. 

Training this model requires ground truth object pose in a video during realistic interaction. We propose a method to synthesize such video dataset in~\cref{subsec:procigen-video}. At training time, we use ground truth human pose and object visibility. At test time, we use PARE~\cite{Kocabas_PARE_2021} to estimate the human pose. For the object, we render the HDM object predictions twice, with and without reconstructed human points, and then compute the mask area ratio as the visibility ratio.

With the predicted rotations, we can then decompose the object tracking into global shape and per-frame object pose optimization. Specifically, we optimize an object shape $\mat{O}\in\mathbb{R}^{N\times 3}$ in canonical space and its corresponding rotation $\mat{R}_i$, translation $\vect{t}_i^o$ and scale $s_i^o$ at frame $i$. We transform the object from canonical space to frame $i$ via: $\mat{O}_i^\prime =s^o_i\mat{O}\mat{R}_i  + \vect{t}_i$. We optimize the shape and pose to fit into 2D object masks $M_i$ and HDM object reconstructions: 
\begin{align}
    &\mathcal{L}_\text{obj} = \sum_i^T \lambda_\text{cd}^o d_\text{cd}(\mat{O}_i^\prime, \mat{P}^o_i) + \lambda_\text{occ} L_\text{occ-sil}(\pi(\mat{O}_i^\prime), M_i) + \notag\\
    &\lambda_a^o L_\text{acc}(\mathcal{O}^\prime) + \lambda_a^r L_\text{acc}(\mathcal{R}) + \lambda_a^t L_\text{acc}(\mathcal{T}) + \lambda_a^s L_\text{acc}(\vect{s})
    \label{eq:loss_object}
\end{align}
where $\pi(\cdot)$ denotes differentiable rendering and $L_\text{occ-sil}$ is the occlusion aware silhouette loss \cite{zhang2020phosa}. $L_\text{acc}$ is temporal smoothness (\cref{eq:loss_human}) for a sequence of object points: $\mathcal{O}^\prime=\{\mat{O}_1^\prime,..., \mat{O}_T^\prime\}$, rotations $\mathcal{R}$, translations $\mathcal{T}$ and scales $\vect{s}$. 

We initialize the canonical object shape $\mat{O}$ from HDM reconstruction of one random frame where the object visibility is higher than a threshold $\sigma$. For more details about loss weights $\lambda_*$, see Supp.

\subsection{Joint human object optimization}\label{subsec:joint-track}
The shape optimization discussed in \cref{subsec:human-recon} and \cref{subsec:object-recon} deals with human and object separately which can lead to unrealistic interaction. 
Hence, we propose to further jointly optimize human and object together to satisfy contacts predicted by the initial HDM~\cite{xie2023template_free} reconstructions. 

HDM reconstructs separate human $\mat{P}^h_i$ and object $\mat{P}^o_i$ point clouds, which allows computing the contact points to pull our optimized human and object together when there are contacts. 
Specifically, we identify the human points that are in contact as the points whose distance to the object is smaller than a threshold: $\Tilde{\mat{P}}_i^h=\{\vect{p}_{i, j}^h | \text{min}_k |\vect{p}_{i, j}^h - \vect{p}_{i, k}^o|_2<\delta, \forall j, k \in \{1, ...,N\}\}$. The corresponding object points $\vect{p}_{i, k}^o$ are identified as the object contacts $\Tilde{\mat{P}}_i^o$. We then transfer the contact points to the optimized human $\mat{H}_i$ and object points $\mat{O}^\prime_i$ by finding their closest points in $\Tilde{\mat{P}}_i^h, \Tilde{\mat{P}}_i^o$. 
Denoting the transferred contact points as $\Tilde{\mat{H}}_i, \Tilde{\mat{O}}_i$, the joint optimization objective combines contact distance loss and separate human and object losses: 
\begin{equation}
    \mathcal{L}_\text{joint} = \mathcal{L}_\text{hum} + \mathcal{L}_\text{obj} + \lambda_c \sum_i ||\Tilde{\mat{H}}_i - \Tilde{\mat{O}}_i||^2_2
    \label{eq:loss_joint-opt}
\end{equation}
where ${L}_\text{hum}$ and ${L}_\text{obj}$ are defined in \cref{eq:loss_human} and \cref{eq:loss_object} respectively. See Supp. for more loss weight details.

\subsection{\dataNameShort{}: Synthetic Interaction Videos}\label{subsec:procigen-video}
Training our video-based \poseName{} requires video datasets of human object interaction. Previous methods\cite{xie22chore, xie2023vistracker} were trained on BEHAVE\cite{bhatnagar22behave} and InterCap~\cite{huang2022intercap} which cannot generalize to other objects as shown in~\cite{xie2023template_free}. Capturing more interaction is not scalable due to the compositionality of human object interaction. Recent work ProciGen~\cite{xie2023template_free} can generate interaction with new objects but is limited to static frames only. To this end, we introduce \dataName{}, a method to synthesize videos for interaction.

We start with ProciGen~\cite{xie2023template_free}, which generates new interactions by replacing the real captured object with new shapes from the same category. To generate an interaction video, we sample a chunk of an interaction sequence with human and object poses from real data. We then randomly select one shape of the same category from shape databases~\cite{shapenet2015, objaverse} and replace the original object with the new shape. We initialize the object pose for the new shape via dense correspondence~\cite{zhou2022art}. This initialization can have interpenetration as the new shape is different from the original, hence, we further fine-tune the human and object pose to satisfy contacts and temporal smoothness. We then render the human with SMPL-D texture~\cite{bhatnagar2019mgn} and object with original textures to obtain images in blender. Please see Supp. for optimization details and example videos.

We apply our method to generate interaction videos of 10 different object categories with interaction motions sampled from BEHAVE and InterCap training set, and object shapes from Shapenet~\cite{shapenet2015}, Objaverse~\cite{objaverse}.  This leads to around 10 hours of video with 8477 sequences and 2M images. We call this dataset \dataName{}, or \dataNameShort{} for short. Since all the objects used in our dataset are aligned in the canonical space, models trained on our data have category-level pose estimation ability and they generalize well to real data (\cref{fig:teaser}, \cref{table:main-result-template-free}, \cref{tab:ablation-procigen-video}). 
We will release our ProciGen-video dataset to facilitate future research.

\begin{figure*}
    \centering
    \includegraphics[width=1.0\linewidth]{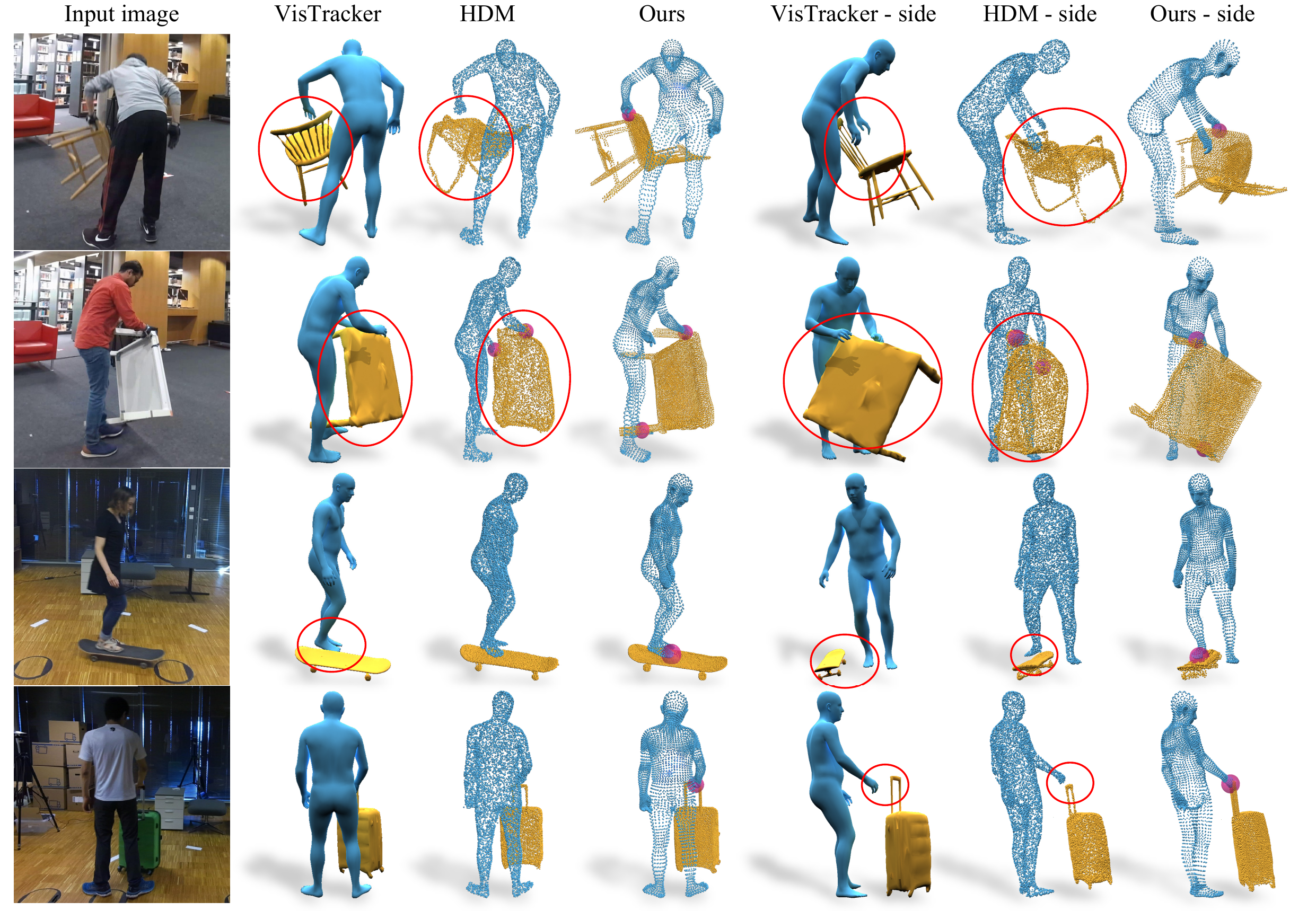}
    \caption{Comparing our method against VisTracker~\cite{xie2023vistracker} and HDM~\cite{xie2023template_free} on BAHEVE (row1-2)~\cite{bhatnagar22behave} and InterCap (row3-4)~\cite{huang2022intercap}. VisTracker relies on post-hoc processing to refine object pose which is inaccurate and HDM reconstructs inconsistent object shapes (row 1-2) or interactions (row 3-4). Our temporal based pose estimation and optimization leads to consistent shape and interaction. }
    \label{fig:comparision-main}
\end{figure*}

\section{Experiments}
In this section, we first compare our method against prior works on reconstructing human object interaction from images or videos and then evaluate our \dataNameShort{} dataset. We further ablate the design choices of our human, object and joint optimization modules. We include network architecture and training details in supplementary. 

\subsection{Comparison with prior methods}\label{subsec:main-comparison}
\textbf{Baselines.} We compare our method against image-based methods CHORE~\cite{xie22chore}, HDM~\cite{xie2023template_free} and video based method VisTracker~\cite{xie2023vistracker}. CHORE relies on known object templates while HDM is template-free. VisTracker also requires known object templates but it improves CHORE on video by leveraging image aligned SMPL and temporal information. HOLD~\cite{fan2024hold} is a template-free method for video based hand-object tracking that is relevant to us. However, it relies on hand-specific inverse skinning and hand-object contact annotations which makes it not directly applicable to our setup. Hence we do not consider HOLD as a baseline. 

\noindent\textbf{Datasets.} We conduct experiments on the BEHAVE~\cite{bhatnagar22behave}, InterCap~\cite{huang2022intercap}, synthetic ProciGen~\cite{xie2023template_free} and our \dataNameShort{} datasets. BEHAVE and InterCap capture realistic humans interacting with 20 and 10 objects, respectively. ProciGen extends these dataset by replacing the objects with shapes from ShapeNet~\cite{shapenet2015}, Objaverse~\cite{objaverse}, and ABO~\cite{collins2022abo} datasets and synthesizing new interaction images. Following HDM~\cite{xie2023template_free}, we first train HDM and our \humAEName{} on the ProciGen dataset and then fine tune HDM on BEHAVE and InterCap training sets. Our \poseName{} is also first trained on \dataNameShort{} and then tuned on BEHAVE and InterCap. We test the methods on sequences from the 10 overlapping categories between \dataNameShort{} and BEHAVE as well as InterCap. In total there are 66 test sequences (51k images) in BEHAVE and 15 sequences (2k images) in InterCap. 

\noindent\textbf{Evaluation Metrics.} Following ~\cite{xie2023template_free, what3d_cvpr19}, we evaluate the shape accuracy using F-score with threshold 0.01m. We report the F-score for human, object and combined shape.  

\begin{table}[ht]
    \centering
    \small
    \begin{tabular}{c| c | c c c }
    \toprule[1.5pt]
         & Method & Human$\uparrow$ & Object$\uparrow$ & Comb.$\uparrow$  \\
         \hline
         \parbox[t]{3mm}{\multirow{4}{*}{\rotatebox[origin=c]{90}{BEHAVE}}} 
         & HDM synth. only  &0.3574& 0.4118 & 0.4118 \\
         & Ours synth. only & 0.3780 & 0.4658 & 0.4488  \\
         & HDM &0.3909& 0.5111 & 0.4622 \\
         & Ours &{\bf 0.4113}& {\bf 0.5849 } & {\bf 0.5169 }\\
        \midrule
        \parbox[t]{3mm}{\multirow{4}{*}{\rotatebox[origin=c]{90}{InterCap}}} 
        & HDM synth. only  & 0.3570 & 0.4342 & 0.4080  \\
        & Ours synth. only & 0.3715 & 0.5141 & 0.4646  \\
         & HDM &0.4325 & 0.6267 & 0.5362 \\
         & Ours & {\bf0.4463}& {\bf 0.6329} & {\bf 0.5555}\\
    \bottomrule[1.5pt]
    \end{tabular}
    \caption{\textbf{Reconstruction results} (F-sc.@0.01m) on BEHAVE~\cite{bhatnagar22behave} and InterCap~\cite{huang2022intercap} of \emph{template-free} methods. Synth. only denotes model trained \emph{only} on synthetic ProciGen\cite{xie2023template_free} and our \dataNameShort{}. Our method consistently outperforms HDM~\cite{xie2023template_free} in all settings.}
    
    \label{table:main-result-template-free}
\end{table}
We first compare with template-free method HDM~\cite{xie2023template_free} in \cref{table:main-result-template-free}. HDM is image-based method that cannot reason temporal consistency and fails when the object is occluded. Following HDM~\cite{xie2023template_free}, we also report the results when both methods are trained only on the synthetic ProciGen and \dataNameShort{} datasets (synth. only). It can be seen that our method consistently improves over HDM in both without and with training on real data. 

We then compare with CHORE~\cite{xie22chore} and VisTracker~\cite{xie2023vistracker} that require object templates. For fair comparison, we adapt our method to use the same object template as the shape $\mat{O}$ and optimize only the transformation parameters. We report the results in \cref{table:main-result-template} following the same Procrustes alignment used by CHORE~\cite{xie22chore} to avoid depth-scale ambiguity. CHORE predicts noisy object poses when they are occluded, leading to inconsistent tracking. VisTracker improves CHORE via ad-hoc pose smoothing and infilling which is still suboptimal. Our method takes temporal information into account in the first place and produces more stable tracking. Notably, our method without template (\cref{table:main-result-template-free}) also archives better performance compared to template-based VisTracker and CHORE. We show some qualitative comparisons of our method without template and baselines in \cref{fig:comparision-main}. It can be seen that our method is more robust to occlusions and produces coherent object shapes. 
\begin{table}[ht]
    \centering
    \begin{tabular}{c| c | c c c }
    \toprule[1.5pt]
         & Method & Human$\uparrow$ & Object$\uparrow$ & Comb.$\uparrow$  \\
         \hline
         \parbox[t]{3mm}{\multirow{3}{*}{\rotatebox[origin=c]{90}{\small BEHAVE}}} & CHORE~\cite{xie22chore} &0.3807& 0.4423 & 0.4224 \\
         & VisTracker~\cite{xie2023vistracker}  & 0.4189 & 0.5607 & 0.5012 \\
         & Ours + template  &{\bf0.4972}& {\bf 0.5854} & {\bf 0.5525} \\
        \midrule
        \parbox[t]{3mm}{\multirow{3}{*}{\rotatebox[origin=c]{90}{\small InterCap}}} & CHORE~\cite{xie22chore} & 0.4065 & 0.4972 & 0.4601 \\
         & VisTracker~\cite{xie2023vistracker}   & 0.4293 & 0.5316 & 0.4902  \\
         & Ours + template & {\bf0.4826}& {\bf 0.5349 } & {\bf 0.5349} \\
    \bottomrule[1.5pt]
    \end{tabular}
    \caption{\textbf{Reconstruction results} (F-sc.@0.01m) on BEHAVE~\cite{bhatnagar22behave} and InterCap~\cite{huang2022intercap} for \emph{template based} methods. Our method obtains more accurate tracking in both datasets. }
    
    \label{table:main-result-template}
\end{table}

\subsection{Evaluating the \dataName{} dataset}\label{subsec:exp-ablate-procigen-video}
Our \dataNameShort{} dataset allows us to train a video-based pose estimator that generalizes to other instances of the same category as all the poses are defined w.r.t an aligned canonical space. To evaluate this, we train our object pose \poseName{} on \dataNameShort{}, BEHAVE, and a combination of both. For other parts of our method, we use the same HDM trained on ProciGen~\cite{xie2023template_free} only and perform the same optimization process. Due to compute limitation, we test only on 14 BEHAVE sequences (9.1k images). We report the object pose error, the chamfer distance between predicted and GT rotation applied to GT meshes, and the final tracking results in \cref{tab:ablation-procigen-video}. It can be seen that the performance gap between models trained on synthetic and real data is small and pre-training model on our \dataNameShort{} can boost the performance. 
\begin{table}[ht]
    \centering
    \footnotesize
    \begin{tabular}{l|>{\centering\arraybackslash}p{1.2cm} >{\centering\arraybackslash}p{0.65cm} >{\centering\arraybackslash}p{0.65cm}>{\centering\arraybackslash}p{0.65cm}}
        \poseName{} Training Data & Obj. CD $\downarrow$ & Hum.$\uparrow$ & Obj.$\uparrow$ & Comb.$\uparrow$ \\
        \hline
        a. \dataNameShort{} only & 4.859 & {\bf 0.3392} & 0.4662 & 0.4327 \\
        b. BEHAVE only & 3.175 & 0.3376 & 0.4815 & 0.4385 \\
        c. \dataNameShort{}+BEHAVE & {\bf 2.922}  & {\bf 0.3392}  & {\bf 0.4980}   & {\bf 0.4468}  \\
    \end{tabular}
    \caption{\textbf{Training object pose \poseName{} on different datasets}. Our \poseName{} trained only on our synthetic \dataNameShort{} generalizes to unseen instances in BEHAVE~\cite{bhatnagar22behave}. Further fine-tuning the model on real data archives the best performance.  }
    \label{tab:ablation-procigen-video}
\end{table}

\subsection{Evaluating object pose estimators}\label{subsec:ablate-obj-pose}
We propose \poseName{}, a video based method to predict category level object rotations conditioned on human pose. We compare our \poseName{} against a prior method CenterPose~\cite{lin2022icra:centerpose} for category-level pose estimation. Note that we exclude methods~\cite{leeTTA6Dpose2023, liDeepIMDeepIterative} that require additional optimization as it is too slow to process video data or methods that require object templates~\cite{nguyenGigaPoseFastRobust, zhangSelfSupervised_category6DOF, weiRGBCategorylevelObject2023} or depth~\cite{cai2024ov9d, wen2021bundletrack, foundationposewen2024}. We train CenterPose, our \poseName{} with and without human pose condition on our synthetic \dataNameShort{} and test it on 14 BEHAVE sequences (same as \cref{subsec:exp-ablate-procigen-video} \cref{tab:ablation-procigen-video}a). Results are shown in \cref{tab:ablation-object-pose}. CenterPose predicts noisy pose under occlusions while our model with human conditioning achieves the best overall results.  

\begin{table}[ht]
    \centering
    \footnotesize
    \begin{tabular}{l| >{\centering\arraybackslash}p{1.15cm} >{\centering\arraybackslash}p{0.65cm} >{\centering\arraybackslash}p{0.65cm}>{\centering\arraybackslash}p{0.65cm} }
        Method & Obj. CD$\downarrow$ & Hum.$\uparrow$ & Obj.$\uparrow$ & Comb.$\uparrow$ \\
        \hline
        a. CenterPose + Our opt. & 7.889 & 0.3332 & 0.3272 & 0.3641 \\
        b. Ours w/o hum. cond. & 4.910 & 0.3380 & 0.4610 & 0.4286 \\
        c. Ours full model & {\bf 4.859} & {\bf 0.3392} & {\bf 0.4662} & {\bf 0.4327} \\
    \end{tabular}
    \caption{\textbf{Object pose error} (Chamfer Distance in cm) and joint tracking results using different object pose estimator. Our temporal based \poseName{} predicts more accurate object poses than CenterPose~\cite{lin2022icra:centerpose} and human condition improves the performance. }
    \label{tab:ablation-object-pose}
\end{table}

\subsection{Evaluating human reconstruction module}\label{subsec:ablate-human-recon}
We propose a simple yet efficient network \humAEName{} to obtain correspondence for a sequence of unordered human points. Keeping other modules the same as \cref{tab:ablation-procigen-video}a, we compare results of using our \humAEName{} or a human registration method NICP~\cite{marin24nicp} for the human reconstruction part in \cref{tab:ablation-human-ae}. It can be seen that our method achieves similar performance but is much faster than NICP. We also report the results of optimizing the latent code of \humAEName{} instead of optimizing SMPL parameters in \cref{tab:ablation-human-ae}b. The latent space entangles human pose and shape which is less controllable than SMPL model~\cite{smpl2015loper}. Optimizing the latent code also leads to unsmooth surface points, please see supplementary for examples and more analysis. Optimizing via the SMPL layer leads to better human reconstruction (\cref{tab:ablation-human-ae}c). 

\begin{table}[ht]
    \centering
    \footnotesize
    \begin{tabular}{l|c c c c}
        Method & Hum.$\uparrow$ & Obj.$\uparrow$ & Comb.$\uparrow$ & Corr. Time$\downarrow$  \\
        \hline
        a. NICP + our opt. &  0.3380 & {\bf 0.4669} &  0.4325 & 72.322 \\
        b. Ours opt. latent & 0.3281  & 0.4631  &  0.4283 & {\bf 0.001} \\
        c. Ours & {\bf 0.3392} & 0.4662 & {\bf 0.4327} & 2.553  \\
    \end{tabular}
    \caption{\textbf{Different methods for human optimization} and the runtime (seconds/image) to obtain correspondences. NICP~\cite{marin24nicp} obtains similar results yet the runtime is significantly longer than our autoencoder (b, c). Optimizing through the SMPL body model achieves a good balance between runtime and accuracy.}
    \label{tab:ablation-human-ae}
\end{table}

\subsection{Importance of different losses}\label{subsec:ablate-losses}
\begin{figure}
    \centering
    \includegraphics[width=1.0\linewidth]{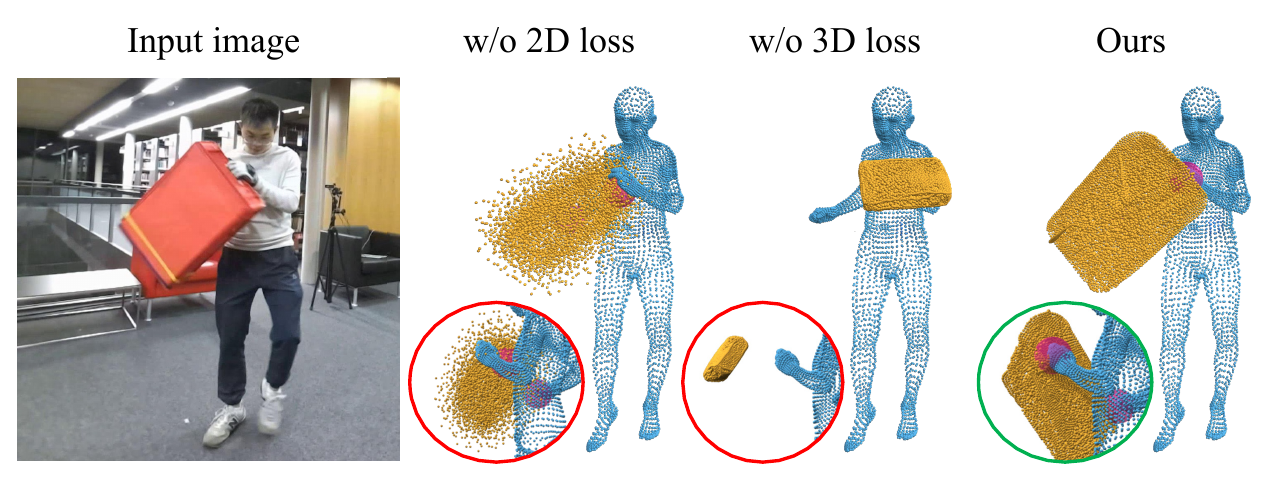}
    \caption{\textbf{The effects of 2D and 3D losses} for object optimization. Without the2D mask loss, the object shape is very noisy and without 3D chamfer loss the relative object position is incorrect.}
    \label{fig:ablate-2d-3d-loss}
\end{figure}

\begin{figure}[ht]
    \centering
    \includegraphics[width=1.0\linewidth]{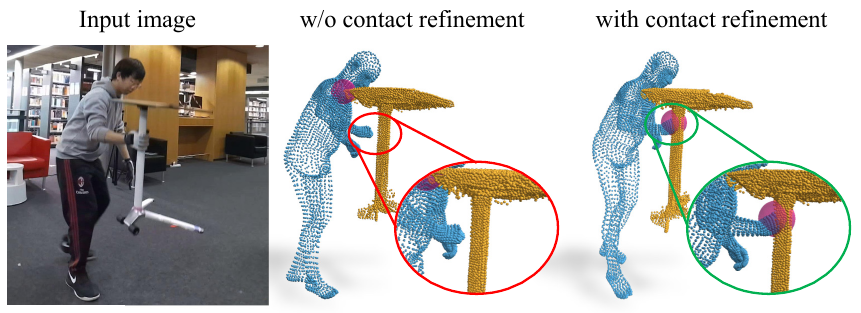}
    \caption{Ablating the influence of the contact-based refinement. Without contact, the hand and object can be far apart, leading to implausible interaction.}
    \label{fig:ablate-contact-loss}
\end{figure}

We ablate the effect of the 2D mask loss ($L_\text{occ-sil}$) and the 3D chamfer loss ($d_\text{cd}$) for object optimization (\cref{eq:loss_object}) in \cref{fig:ablate-2d-3d-loss}. It can be seen that omitting both will lead to a low-fidelity object shape. We also ablate our contact-based pose refinement in \cref{tab:ablation-losses} and Fig.~\ref{fig:ablate-contact-loss}. Quantitatively, the contact refinement has tiny difference yet qualitatively it significantly improves the physical plausibility of the reconstructed interaction (\cref{fig:ablate-contact-loss}). Without this refinement, the object will be floating in the air which is not physically plausible. 

\begin{table}[ht]
    \centering
    \small
    \begin{tabular}{l|c c c}
        Method & Hum. & Obj. & Comb. \\
        \hline
        a. w/o contact refinement & {\bf 0.3430} & 0.4653  & {\bf 0.4338} \\
        b. Ours full model  & 0.3392 & {\bf 0.4662} & 0.4327  \\
    \end{tabular}
    \caption{\textbf{Contact based refinement} leads to tiny quantitative difference but more plausible interaction qualitatively, see \cref{fig:ablate-contact-loss}. }
    \label{tab:ablation-losses}
\end{table}

\section{Limitations and Future Works}
Despite impressive performance on benchmark datasets and strong generalization to real videos, there are still some limitations of our method. First, our method does not reconstruct the textures of the human and object. Our method is easily compatible with Gaussian Splating~\cite{kerbl3Dgaussians} and adding colors to each point could potentially further constraint the optimization~\cite{das2023npg}. Second, our dataset contains only the categories from BEHAVE and InterCap. Future works can capture more objects or explore synthesizing new interactions without real data. Multi-human, multi-object interaction with deformable object tracking are also interesting directions to explore. We leave these for future works. We also provide example failure cases and analysis in Supp.

\section{Conclusion}
In this paper, we present \methodName{}, the first approach to track full-body interaction with dynamic object from monocular RGB video without pre-scanned object templates. Our first contribution is a simple yet efficient autoencoder \humAEName{} to directly predict SMPL vertices from unordered human reconstructions. Our second contribution is a video-based pose estimator \poseName{} that leverages temporal information to predict smooth rotation under occlusions. We also introduce a method to generate synthetic videos for interaction and we synthesize 10-hour videos of $\sim$8.5k sequences for training video based methods. 

Our experiments on BEHAVE and InterCap show that our method significantly outperforms prior template-free and template-based approaches. Ablation studies also show that our method trained on our synthetic video dataset generalizes to real data. We also show that our \humAEName{} is much more efficient than SoTA human registration method with similar performance and our \poseName{} is more accurate than another category-level pose estimator method. Our code, models and dataset will be publicly released.

\noindent
{\footnotesize
\textbf{Acknowledgements.} We thank RVH group members \cite{rvh_grp} for their helpful discussions. This work is funded by the Deutsche Forschungsgemeinschaft (DFG, German Research Foundation) - 409792180 (Emmy Noether Programme,
project: Real Virtual Humans), and German Federal Ministry of Education and Research (BMBF): Tübingen AI Center, FKZ: 01IS18039A, and Amazon-MPI science hub. Gerard Pons-Moll is a Professor at the University of Tübingen endowed by the Carl Zeiss Foundation, at the Department of Computer Science and a member of the Machine Learning Cluster of Excellence, EXC number 2064/1 – Project number 390727645.
}
{
    \small
    \bibliographystyle{ieeenat_fullname}
    \bibliography{main}
}
\clearpage
\setcounter{page}{1}

\maketitle
%\maketitlesupplementary

In this supplementary, we provide more implementation details for our tracking method and synthetic video generation. We also further analyze the design considerations and discuss typical failure cases of our method. Please refer to our supplementary video for video tracking results. 

\section{Implementation Details}
We discuss the network architecture, training, and optimization details in this section. Our code will be publicly released with clear documentation to foster future research. 

\subsection{\humAEName{} and human optimization}
For our human \humAEName{}, we adapt the encoder from PVCNN~\cite{liu2019pvcnn} which is also used in \cite{melaskyriazi2023pc2, xie2023template_free}. It compresses input point clouds of shape $N\times 3$ into downsampled point feature of shape $512\times 16$. We add two additional point convolution layer~\cite{liu2019pvcnn} to further compress it into latent vector $\vect{z}^h$ of shape $1024\times 1$. The latent code is then sent to one MLP layer, followed by 6 blocks of MLP layers with residual connection. The MLP compress the latent code to 512 dimension and each block consists of three MLP layers with LeakeyReLU activation. The output dimensions of the MLPs in each block are $256, 256, 512$. The 512 dimension feature vector is then sent to a large MLP which predicts 6890 SMPL vertices as a single vector. 

We train our \humAEName{} with a loss weight $\lambda_\text{v2v}=100$ for the vertex to vertex loss and use Adam optimizer with learning rate of 3e-4. The model is trained on the GT SMPL meshes from ProciGen training set. It takes around 12 hours to finish training on 4 RTX8000 GPUs with batch size 32. The loss weights for the human optimization are: $\lambda_\text{cd}^h=100, \lambda_p=1e-5, \lambda_\text{acc}=100$. We use Adam of learning rate 0.001 and stochastic gradient descent to optimize the human pose parameters, with a batch size of 256. We optimize for 2500 steps which takes $\sim$30 minutes on an A40@40GB GPU.

\subsection{\poseName{} and oject optimization}
For the object pose \poseName{}, we combine DINOv2\cite{oquab2024dinov2} image encoder with transformer~\cite{NIPS2017_attention}. DINOv2 encodes image of shape $3\times 224\times 224$ into a feature grid of $768\times 16\times 16$. We then add three 2D convolution layers with kernel size 4, group normalization and leaky ReLU activation to further compress the feature grid into a vector of shape $1\times1\times 768$. This operation is similar to the one used in MagicPony~\cite{wu2023magicpony}. The dimension of human feature is $294=25\times6+24\times6$, which consists of 25 body joints and their velocities and SMPL body pose represented as rotation 6D\cite{Zhou_2019_CVPR_rot6d}. We encode the human feature using two MLPs with a latent dimension 128 and output dimension 128. The human feature is then concatenated with object visibility and image feature vector and sent to transformer with 3 encoder layers~\cite{NIPS2017_attention}. Each encoder layer has 4 heads and feed forward dimension of 256. The temporal features are then sent to 3 MLP layers with output dimensions of 128, 64, and 6. 

We train the model with learning rate 3e-4 (Adam optimizer) and batch size 16, temporal window size 16. It takes around 31 hours to converge on 4 RTX8000 GPUs. We train two models for all 10 categories in \dataNameShort{} dataset: one for large objects (chair, table, monitor) and another one for small symmetric objects (all the rest categories). The loss weights for the object optimization are: $\lambda^o_\text{cd}=10, \lambda_\text{occ}=0.001, \lambda_a^r=1000, \lambda_a^t=200, \lambda_a^s=1000$. We optimize canonical shape and per-frame poses with a batch size of 64. For models trained on synthetic data only we optimize for 16k steps as the initial shape is less accurate, which takes around 2 hours. For models fine-tuned on real data, we optimize only 6k steps which takes 50-60 minutes on one A40 GPU. 

\subsection{Joint optimization}
The loss weights for the human ($\mathcal{L}_\text{hum}$) and object ($\mathcal{L}_\text{obj}$) loss terms are the same as the ones used for separate optimization. The contact loss weight $\lambda_c=10$. Note that we optimize only the SMPL body pose and object rotation parameters as this is used only for fine tuning the poses. 

Similar to separate optimization, we use Adam with learning rate 0.001 for human and 6e-4 for object. We refine for 2500 steps with batch size 64, which takes in total $\sim$35 minutes on one A40 GPU.

\subsection{\dataName{} data generation}
\begin{figure*}[th]
    \centering
    \includegraphics[width=1.0\linewidth]{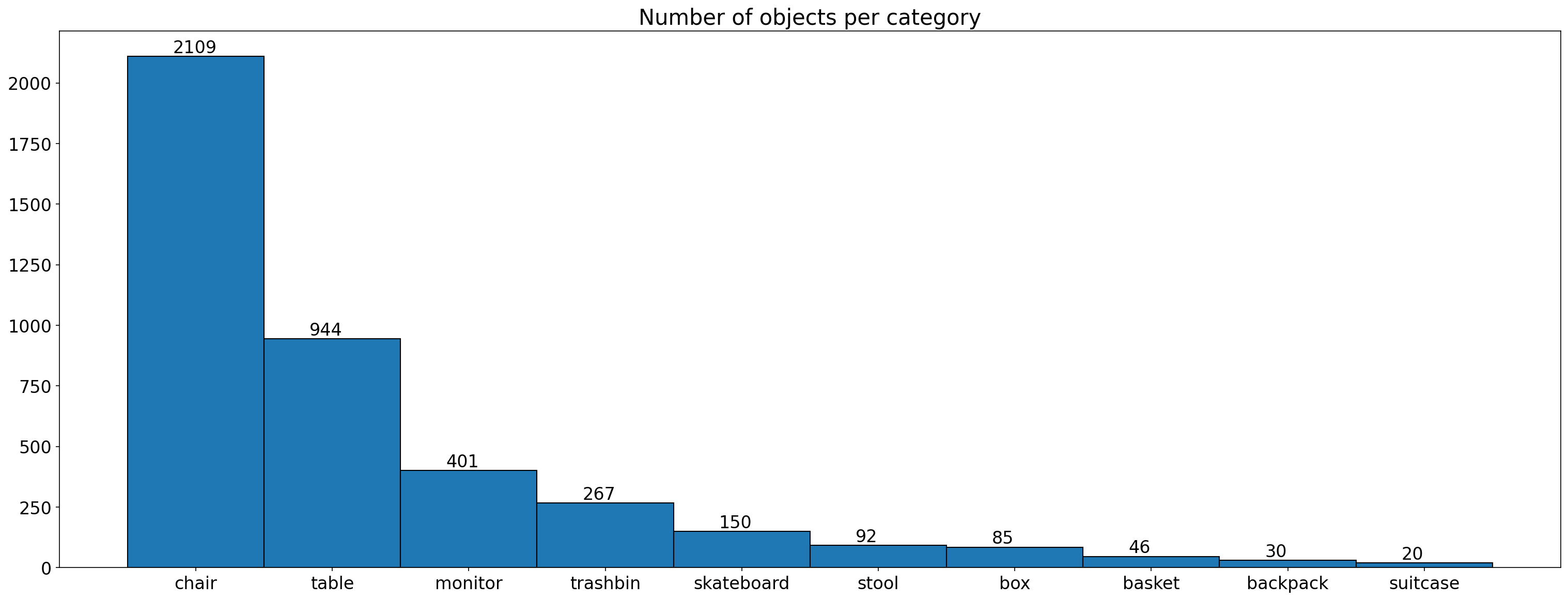}
    \caption{\textbf{Number of distinct object shapes} used in our \dataNameShort{} dataset. Our method is scalable and can generate interaction for new object shapes within these categories. }
    \label{fig:procigen-video-shapes}
\end{figure*}
\begin{figure*}[th]
    \centering
    \includegraphics[width=1.0\linewidth]{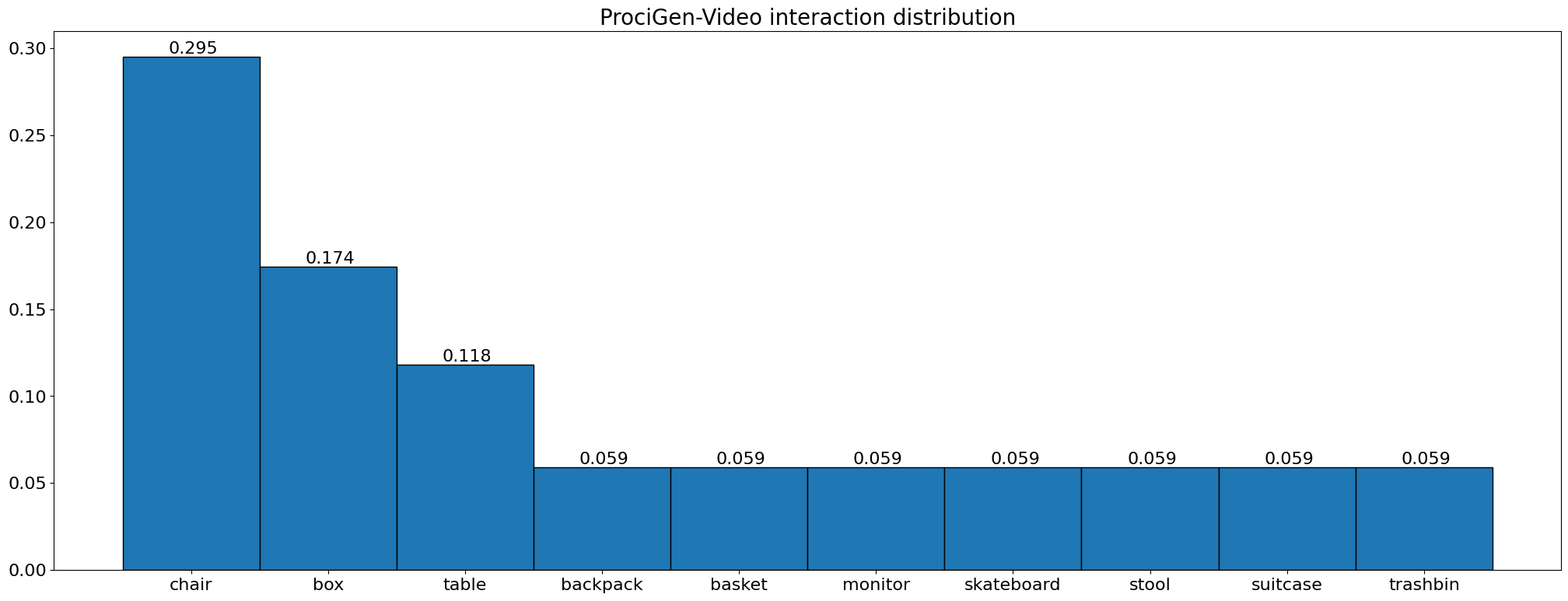}
    \caption{\textbf{Distribution of interaction sequences per category} in our \dataNameShort{} dataset. Our dataset is balanced for most categories except for chair which contains more complex shape and interactions.}
    \label{fig:procigen-video-seqs}
\end{figure*}
We start from ProciGen proposed in \cite{xie2023template_free} to procedurally generate interaction videos for new object shapes. 
The goal is to change the human and object shape and render new videos. We first sample a chunk of human and object poses from interaction sequences in real data. The human is represented using SMPL~\cite{smpl2015loper} pose $\Theta=\{\pose_1,...,\pose_N\}$ and shape $\mathcal{B}=\{\beta_1,...\beta_N\}$ parameters, here $1,...,N$  are the time index. We compute dense correspondence between original object shape and new shape using an autoencoder~\cite{zhou2022art}, which allows transferring contacts from original shape to new shape. We also use the correspondence to initialize the pose $\mathbf{T}_i\in\mathbb{R}^{4\times 4}$ for the new object~\cite{xie2023template_free}. The initialization can lead to interpenetration problem, hence we further optimize the body poses $\Theta$, shapes $\mathcal{B}$ and object transformations $\mathcal{T}=\{\mathbf{T}_1,...\mathbf{T}_N\}$ to satisfy contacts and temporal smoothness: 
\begin{equation}
    \mathcal{L}(\Theta, \mathcal{T}, \mathcal{B}) = \lambda_c L_c + \lambda_n L_n + \lambda_\text{colli} L_\text{colli} + \lambda_\text{init} L_\text{init} + \lambda_\text{acc} L_\text{acc}
\end{equation}
where the contact loss $L_c$, normal loss $L_n$, interpenetration $L_\text{init}$ and initialization penalty $L_\text{init}$ are defined in \cite{xie2023template_free}. And $L_\text{acc}$ is the temporal smoothness loss defined in \cref{eq:loss_human} applied to a sequence of SMPL vertices. Note that we also randomly sample a body shape parameter from the MGN dataset~\cite{bhatnagar2019mgn} to replace the original shape for more diversity. The loss weights used are: $L_c=400, L_n=6.25, L_\text{colli}=9, L_\text{init}=100, L_\text{acc}=10$. 

Once optimized, we use SMPL-D registration~\cite{bhatnagar2019mgn} which adds per-vertex offsets to the SMPL vertices to model clothing deformation and texture. For the object, we use the original textures from the CAD model. We also add small random global rotation and translation to the full sequence to increase diversity. We render the human and object in blender with random lighting and no backgrounds. Some example renderings can be found in ADD REF. 

We generate interaction videos for 10 object categories. The interaction poses are sampled from BEHAVE~\cite{bhatnagar22behave} and InterCap~\cite{huang2022intercap}, object shapes are sampled from Objaverse~\cite{objaverse} and ShapeNet~\cite{shapenet2015}. The distribution of distinct object shapes can be found in\cref{fig:procigen-video-shapes}, and the number of interaction sequences per-category can be found in \cref{fig:procigen-video-seqs}.
The original BEHAVE and ShapeNet are captured at 30fps, we generate synthetic data at 15fps and each sequence has 64 frames (4.27 seconds). In total, we generate 8477 sequences which amounts to 10 hours long videos. Our method can scale up to include more objects and longer videos, which is much more scalable than capturing real data.

\section{Additional Analysis and Result}
In this section, we provide additional analysis to the design considerations of our human and object reconstruction modeuls. We also show generalization to unseen category. Please refer to our video for more results and comparison.

\subsection{Object pose \poseName{}}
\begin{figure}
    \centering
    \includegraphics[width=1.0\linewidth]{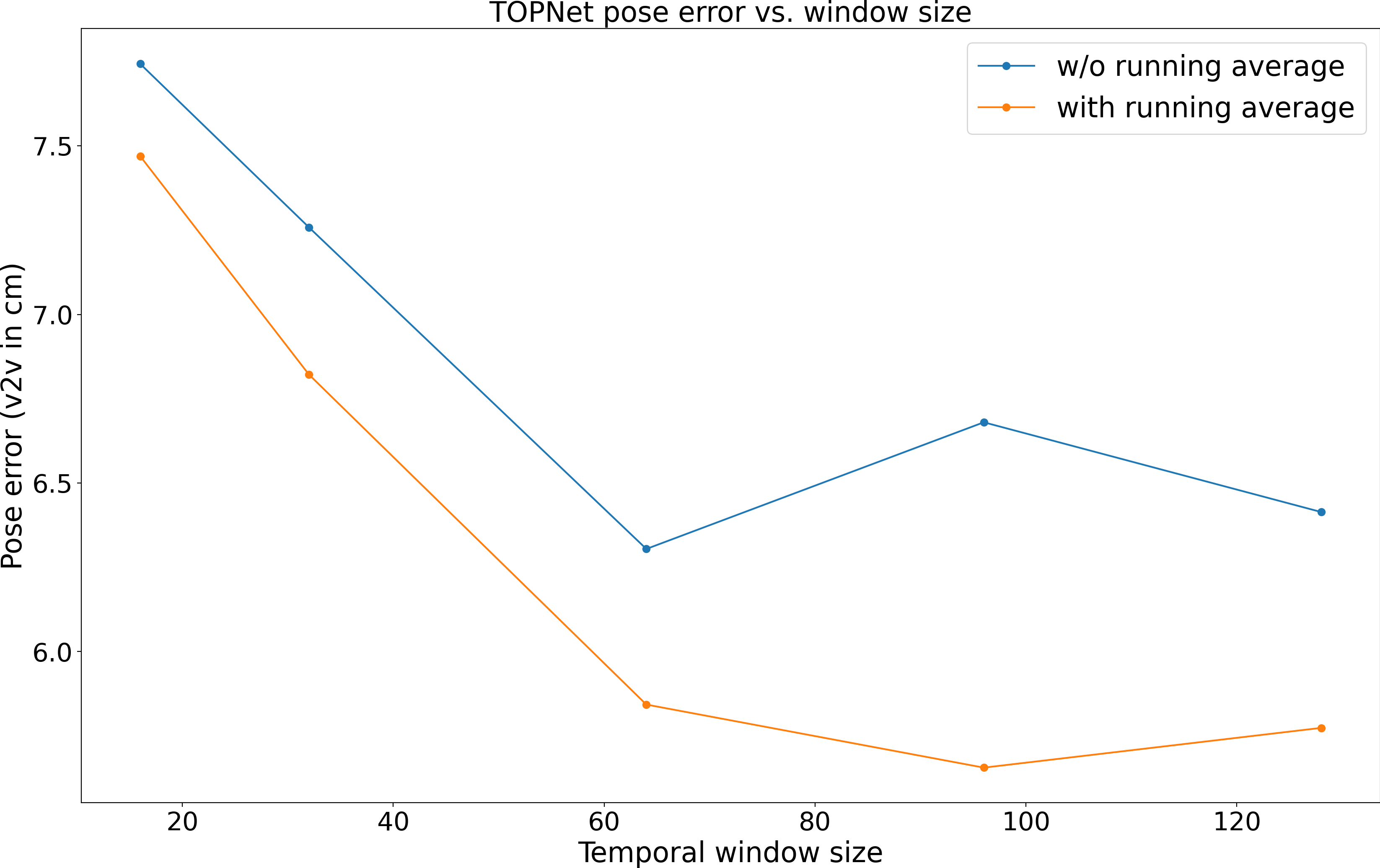}
    \caption{\textbf{Object pose error versus the temporal window size} used at inference time. The model was trained with window size=16. Averaging predictions of each frame in different sliding windows consistently leads to better pose estimations.}
    \label{fig:supp-topnet-pose-errors}
\end{figure}

Our \poseName{} computes cross attention between W consecutive images and directly predicts W rotations for them. We train our model with $W=16$ due to limited IO speed: with a batch size of 16, it needs to load 256 images with corresponding GT data which already takes $0.6\sim1$ second. Using longer window size significantly increases the training time. In contrast, we find that the learned attention weights can be applied larger window size even though the model is trained for $W=16$ only. We plot the object pose error with different test time window size in \cref{fig:supp-topnet-pose-errors}. Here we report the pose error as the vertex to vertex error (cm) after applying predicted and GT rotation to the GT object vertices. We apply a sliding window of size W to process the full sequence, which means each image can appear several times at different sliding windows. We average predictions of all possible sliding windows, which also leads to smoother and more accurate pose, see \cref{fig:supp-topnet-pose-errors} (with running average).

\subsection{Human Reconstruction}
\begin{figure}
    \centering
    \includegraphics[width=1.0\linewidth]{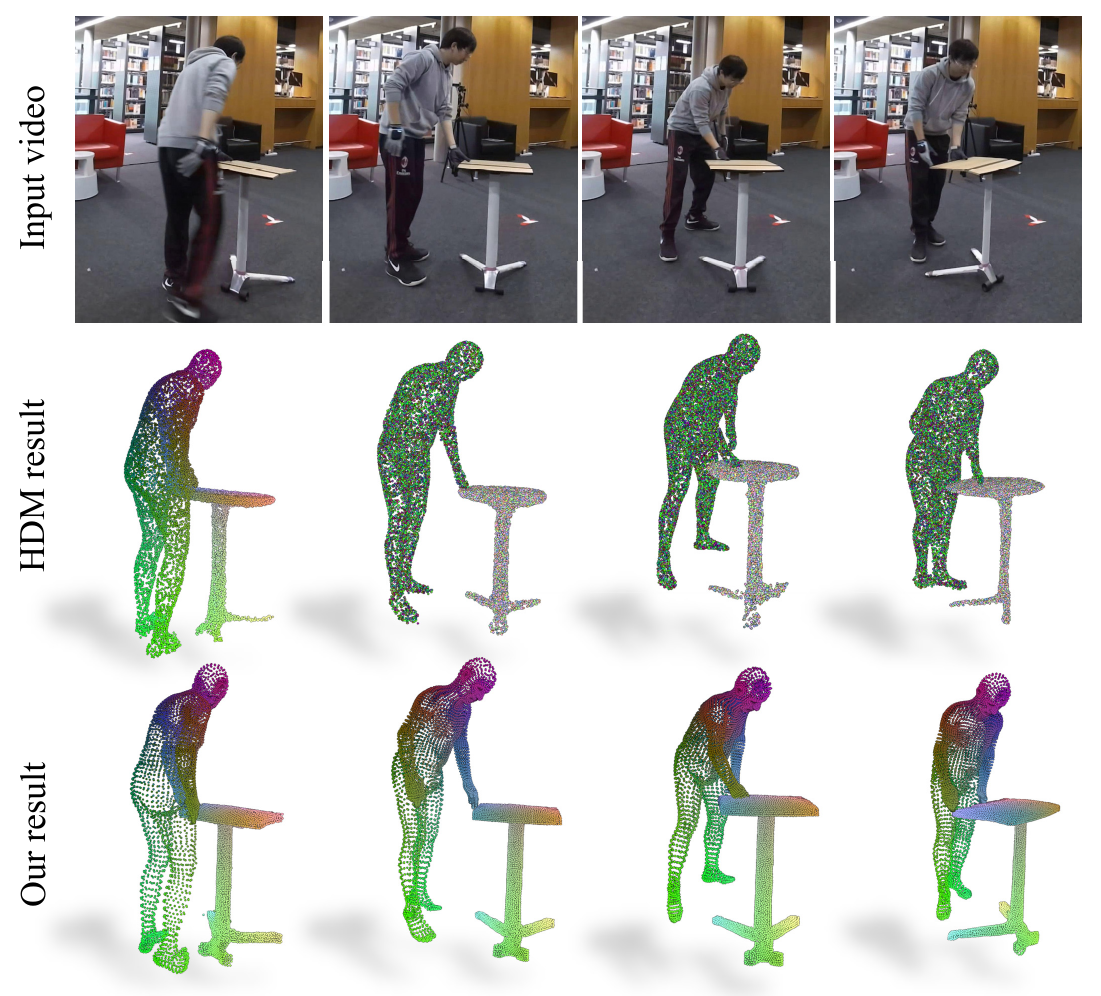}
    \caption{\textbf{Visualization of the correspondence.} HDM~\cite{xie2023template_free} outputs unordered points while our method consistently tracks the human and object across frames.}
    \label{fig:supp-viz-corr}
\end{figure}
We compare the correspondence across frames from HDM and our method in \cref{fig:supp-viz-corr}. HDM is image-based method and outputs point clouds without any ordering. On the other hand, our method tracks the point across the full sequence.

\begin{figure}
    \centering
    \includegraphics[width=1.0\linewidth]{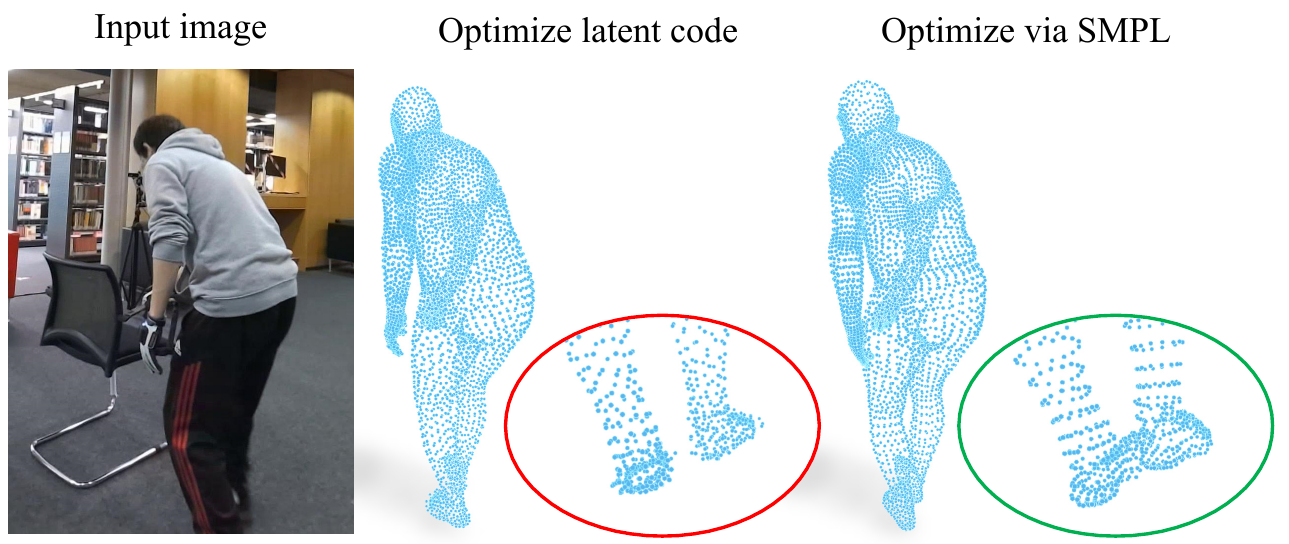}
    \caption{\textbf{The problem of optimizing \humAEName{} latent code}. The latent space of our \humAEName{} entangles human pose and shape. Optimizing it directly also leads to less smooth surface.}
    \label{fig:supp-opt-human-latent}
\end{figure}

We argue in \cref{subsec:human-recon} that the latent space of our \humAEName{} is less interpretable which leads to slightly worse result compared to optimizing via SMPL layer (\cref{tab:ablation-human-ae}). Here we visualize another problem of optimization via the \humAEName{}: the surface points become less smooth, see \cref{fig:supp-opt-human-latent}. It can be seen that some points on the feet spread out from the original position, leading to a noisy surface. In contrast, optimizing via SMPL layer guarantees a smooth surface.

\subsection{Generalization to unseen categories}
\begin{figure*}[th]
    \centering
    \includegraphics[width=1.0\linewidth]{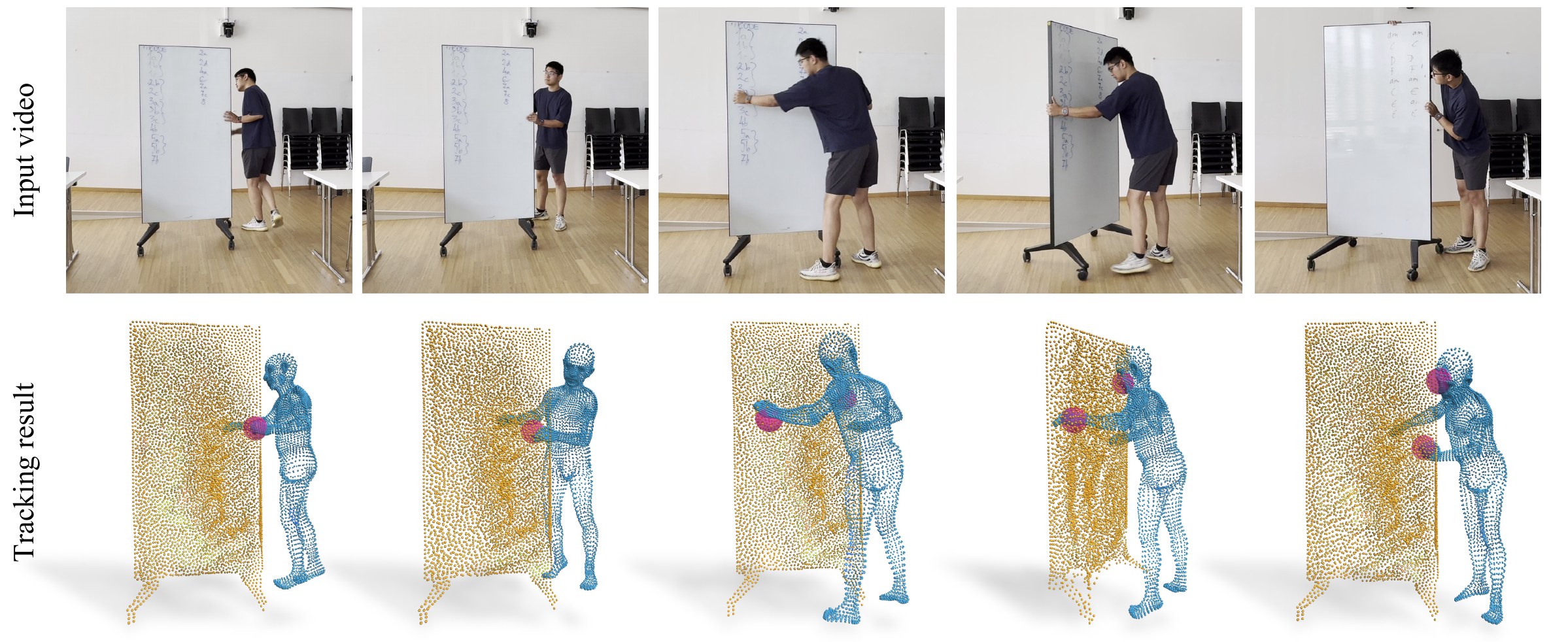}
    \caption{\textbf{Generalization to unseen category.} We test our method to unseen category blackboard. It can be seen that our method can reconstruct the shape and tracks the human object interaction.}
    \label{fig:supp-ood-generalization}
\end{figure*}
Our model was trained on ten common daily life object categories. It works well for new object instances of the same category, as can be seen in \cref{fig:teaser} and our supplementary video. We also test our method on unseen category in \cref{fig:supp-ood-generalization}. In general, our method can work on new categories that are similar to those seen in our training set.

\section{Failure Case Analysis}
\begin{figure*}[th]
    \centering
    \includegraphics[width=1.0\linewidth]{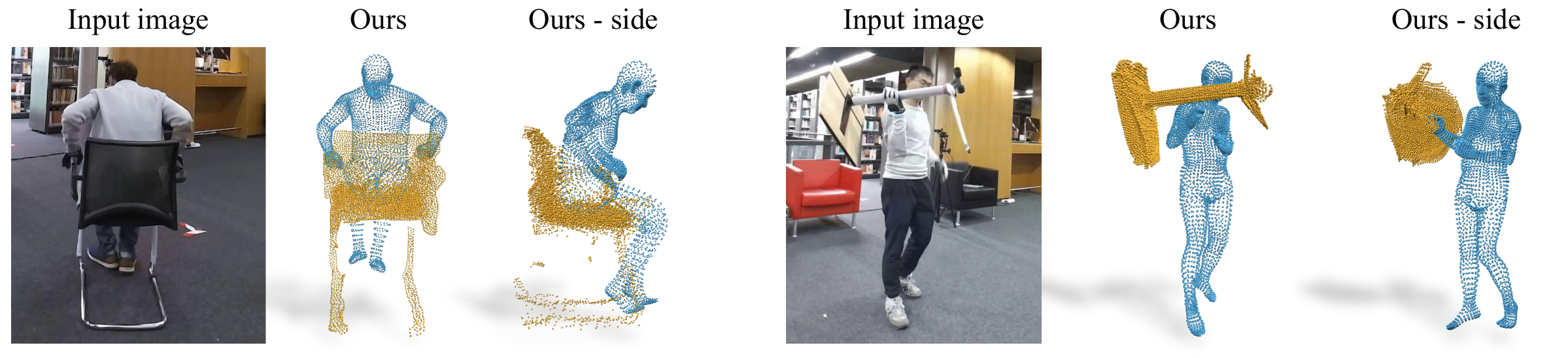}
    \caption{\textbf{Example failure cases.} Our method fails to reconstruct the object shape (left) as only one view of the object is seen in the entire video. It can also struggle to predict extreme rare pose (right), leading to less faithful shape and tracking.}
    \label{fig:supp-failure}
\end{figure*}
We show two typical failure cases of our method in \cref{fig:supp-failure}. Overall, our method tracks humans reliably in most cases while object tracking is more challenging due to occlusions and lack of template shapes. Our method can produce noisy object shape when there are not enough views to reconstruct the object. In \cref{fig:supp-failure} left, the chair remains static in the full sequence, hence our method only receives information about the chair in back side view. The object shape aligns well with the input image but the 3D structure is not coherent. Future works can further improve our method by imposing stronger object shape prior. For example, optimizing via a well-behaved latent space which provides better output shape. 

Our method can also predict noisy object pose under rare interaction like \cref{fig:supp-failure} right. In this sequence, noisy poses dominate the optimization, leading to inaccurate shape and tracking. Training on more object poses or with additional data augmentation is a possible direction to explore. Foundationpose~\cite{foundationposewen2024} trained their model on millions of different objaverse~\cite{objaverse} objects hence has better generalization. However, they rely on CAD model and depth input. One interesting direction is to develop methods that can iteratively improve object shape reconstruction and pose estimation. With our \poseName{}, one can obtain initial object reconstruction, which should be helpful to improve object pose estimation. This iterative mutual improvement should lead to better shape and pose tracking. 

Further more, our method does not deal with object symmetries explicitly. Future works can adopt good practices from object pose estimation community~\cite{Wang_2021_GDRN, di_so-pose_2021, schoenberger2016sfm_COLMAP} to further enhance the robustness of our method. 

\begin{figure*}[t]
    \centering
    \includegraphics[width=1.0\linewidth]{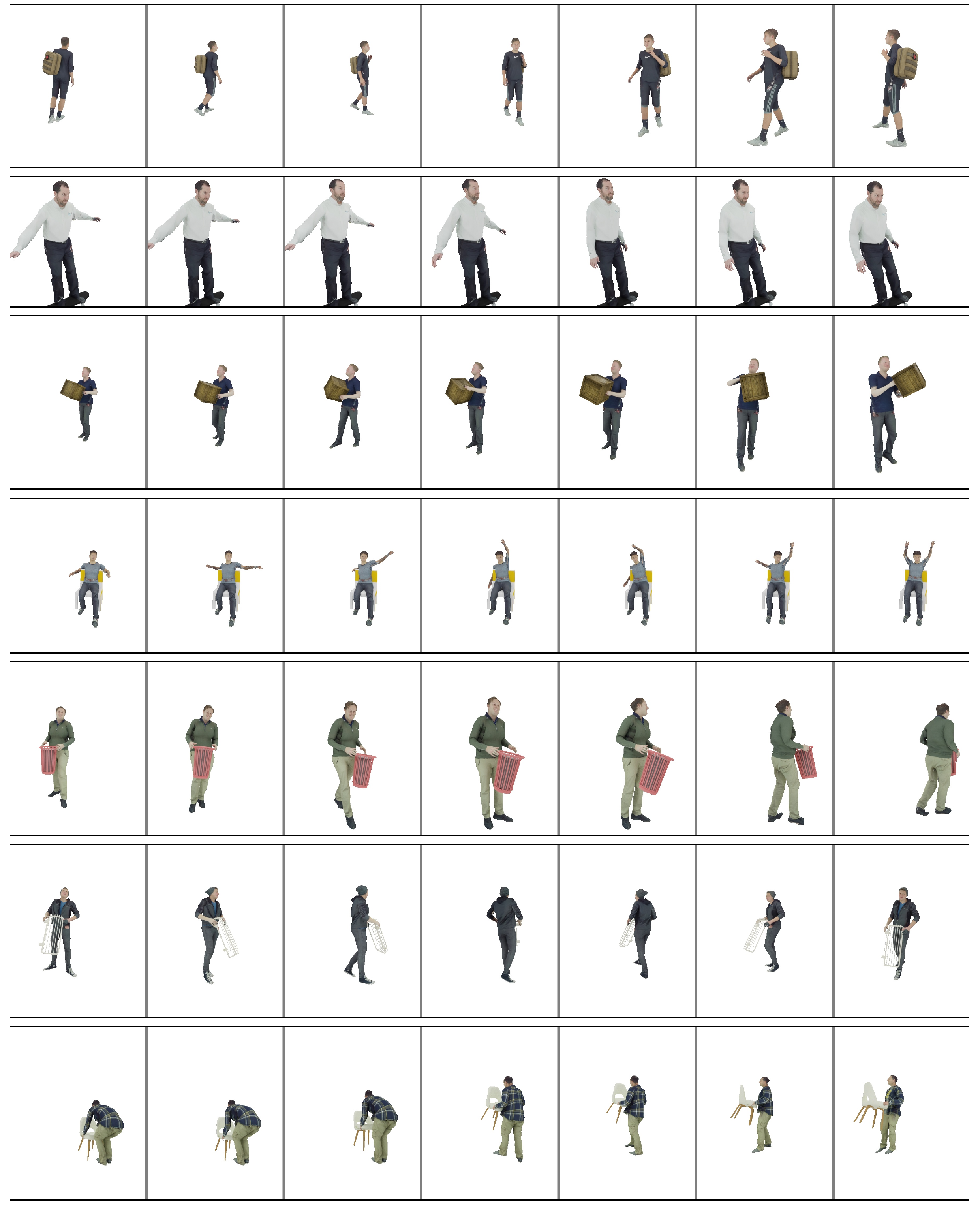}
    \caption{\textbf{Example sequences from our \dataName{} dataset.} We generate realistic interactions with diverse object shapes. Please refer to our supplementary video for more examples.}
    \label{fig:procigen-video-examples}
\end{figure*}

\end{document}